\documentclass{llncs}
\usepackage{makeidx}  

\usepackage{verbatim}
\usepackage{amsfonts}
\usepackage{nccmath}
\usepackage[algo2e,ruled,vlined]{algorithm2e}
\usepackage{caption}
\usepackage{subfigure}
\usepackage{stackengine}
\usepackage{multirow}

\usepackage{epstopdf}

\usepackage{floatrow}

\usepackage{xcolor}
\usepackage[export]{adjustbox}

\usepackage{booktabs}

\usepackage{tabularx}

\newcolumntype{L}[1]{>{\raggedright\let\newline\\\arraybackslash\hspace{0pt}}m{#1}}
\newcolumntype{C}[1]{>{\centering\let\newline\\\arraybackslash\hspace{0pt}}m{#1}}
\newcolumntype{R}[1]{>{\raggedleft\let\newline\\\arraybackslash\hspace{0pt}}m{#1}}

\newfloatcommand{capbtabbox}{table}[][\FBwidth]

\renewcommand{\vec}[1]{\boldsymbol{#1}}

\newcommand{\mr}[1]{\mathrm{#1}}
\newcommand{\tx}[1]{\textrm{#1}}
\newcommand{\tr}[1]{#1^\top}

\newcommand{\RR}{\mathbb{R}}

\newcommand{\AAA}{A}
\newcommand{\QQ}{Q}
\newcommand{\WW}{W}
\newcommand{\DD}{D}

\newcommand{\ww}{\vec w}

\DeclareMathOperator*{\argmin}{arg\,min}

\newcommand{\beq}{\begin{equation}}
	\newcommand{\eeq}{\end{equation}}
\newcommand{\beqa}{\begin{eqnarray}}
	\newcommand{\eeqa}{\end{eqnarray}}
\newcommand{\beqas}{\begin{eqnarray*}}
	\newcommand{\eeqas}{\end{eqnarray*}}

\expandafter\def\expandafter\normalsize\expandafter{%
	\normalsize
	\setlength\abovedisplayskip{6pt}
	\setlength\belowdisplayskip{6pt}
	\setlength\abovedisplayshortskip{6pt}
	\setlength\belowdisplayshortskip{6pt}
}

\DeclareMathSizes{10}{10}{6}{5}

\title{\Large White matter fiber segmentation using functional varifolds}

\author{Kuldeep Kumar$^{1,2}$, Pietro Gori$^{2,4}$, Benjamin Charlier$^{2,5}$, Stanley Durrleman$^{2}$,\\ Olivier Colliot$^{2,3}$ \and Christian Desrosiers$^{1}$}

\institute{$^{1}$ LIVIA, \'Ecole de technologie sup\'erieure (\'ETS), Montreal, Canada \\
	$^{2}$ Aramis project-team, Inria Paris, Sorbonne Universit\'es, UPMC Univ Paris 06, Inserm, CNRS, Institut du cerveau et la moelle (ICM) - H\^opital Piti\'e-Salp\^etri\`ere, Boulevard de l$\rq$h\^opital, F-75013, Paris, France \\
	$^{3}$ AP-HP, Departments of Neurology and Neuroradiology, H\^opital Piti\'e-Salp\^etri\`ere, 75013, Paris, France  \\
	$^{4}$ LTCI lab - IMAGES group, T\'el\'ecom ParisTech, Paris, France \\
	$^{5}$ Universit\'e de Montpellier, France}

\begin{document}
	
	\maketitle
	
	\begin{abstract}
		The extraction of fibers from dMRI data typically produces a large number of fibers, it is common to group fibers into bundles. To this end, many specialized distance measures, such as MCP, have been used for fiber similarity. However, these distance based approaches require point-wise correspondence and focus only on the geometry of the fibers. Recent publications have highlighted that using microstructure measures along fibers improves tractography analysis. Also, many neurodegenerative diseases impacting white matter require the study of microstructure measures as well as the white matter geometry. Motivated by these, we propose to use a novel computational model for fibers, called functional varifolds, characterized by a metric that considers both the geometry and microstructure measure (e.g. GFA) along the fiber pathway. We use it to cluster fibers with a dictionary learning and sparse coding-based framework, and present a preliminary analysis using HCP data.
	\end{abstract}
	
	\section{Introduction}
	\label{Introduction}
	
	Recent advances in diffusion magnetic resonance imaging (dMRI) analysis have led to the development of powerful techniques for the non-invasive investigation of white matter connectivity in the human brain. By measuring the diffusion of water molecules along white matter fibers, dMRI can help identify connection pathways in the brain and better understand neurological diseases related to white matter \cite{hagmann2006understanding}. Since the extraction of fibers from dMRI data, known as tractography, typically produces a large number of fibers, it is common to group these fibers into larger clusters called \emph{bundles}. Clustering fibers is also essential for the creation of white matter atlases, visualization, and statistical analysis of microstructure measures along tracts \cite{o2009tract}.
	
	Most fiber clustering methods use specialized distance measures, such as Mean Closest Points (MCP) distance \cite{corouge2004towards,moberts2005evaluation}. However, these distance-based approaches require point-wise correspondence between fibers and only consider fiber geometry. Another important aspect for white matter characterization is the statistical analysis of microstructure measures. As highlighted in recent publications, using microstructure measures along fibers improves tractographic analysis \cite{colby2012along,maddah2008unified,o2009tract,wang2013application,wassermann2010unsupervised,yeatman2012tract}. Motivated by these, we propose to use a novel computational model for fibers, called functional varifolds, characterized by a metric that considers both the geometry and microstructure measure (e.g. generalized fractional anisotropy) along fiber pathways.
	
	Motivation for this work comes from the fact that the integrity of white matter is an important factor underlying many cognitive and neurological disorders. In vivo, tissue properties may vary along each tract for several reasons: different populations of axons enter and exit the tract, and disease can strike at local positions within the tract. Hence, understanding diffusion measures along each fiber tract (i.e., tract profile) may reveal new insights into white matter organization, function, and disease that are not obvious from mean measures of that tract or from the tract geometry alone \cite{colby2012along,yeatman2012tract}. Recently, many approaches have been proposed for tract based morphometry \cite{o2009tract}, which perform statistical analysis of microstructure measures along major tracts after establishing fiber correspondences. While studies highlight the importance of microstructure measures, most approaches either consider the geometry or signal along tracts, but not both. The intuitive approach would be to consider microstructure signal during clustering also. However, this has been elusive due to lack of appropriate framework.  
	
	As a potential solution, we explore a novel computational model for fibers, called functional varifolds \cite{charlier2014fshape}, which is a generalization of the varifolds framework \cite{charon2013varifold}. The advantages of using functional varifolds are as follows. First, functional varifolds can model the fiber geometry as well as signal along the fibers. Also, it does not require pointwise correspondences between fibers. Lastly, fibers do not need to have the same orientation as in the framework of currents \cite{gori2014prototype}. We test the impact of this new computational model on a fiber clustering task, and compare its performance against existing approaches for this task. 
	
	As clustering method, we reformulate the dictionary learning and sparse coding based framework proposed in \cite{kuldeepMiccai2015,kumar2016sparse,KUMAR2017242}. This choice of framework is driven by its ability to describe the entire data-set of fibers in a compact dictionary of prototypes. Bundles are encoded as sparse non-negative combinations of multiple dictionary prototypes. This alleviates the need for explicit representation of a bundle centroid, which may not be defined or may not represent an actual object. Also, sparse coding allows assigning single fibers to multiple bundles, thus providing a soft clustering. 
	
	The contributions of this paper are threefold: 1) a novel computational model for modeling both fiber geometry and signal along fibers, 2) a generalized clustering framework, based on dictionary learning and sparse coding, adapted to the computational models, and 3) a comprehensive comparison of fully-unsupervised models for clustering fibers. 
	
	\vspace{-3mm}
	\section{White matter fiber segmentation using functional varifolds}
	\label{sec:Coding_And_DL}
	
	\vspace{-3mm}
	\subsection{Modeling fibers using functional varifolds}
	
	In the framework of functional varifolds \cite{charlier2014fshape,charon2013varifold}, a fiber $X$ is assumed to be a polygonal line of $P$ segments described by their center point $x_p \in \RR^3$ and tangent vector $\vec{\beta_p} \in \RR^3$ centered at $x_p$ and of length $c_p$ (respectively, $y_q \in \RR^3$, $\vec{\gamma_q} \in \RR^3$ and $d_q$ for a fiber $Y$ with $Q$ segments). Let $f_p$ and $g_p$ be the signal values at center points $x_p$ and $y_q$ respectively, and $\omega$ the vector field belonging to a reproducing kernel Hilbert space
	(RKHS) $W^*$. Then the fibers $X$ and $Y$ can be modeled based on functional varifolds as: $V_{(X,f)}(\omega) \approx \sum_{p=1}^{P} \omega(x_p,\vec{\beta_p},f_p)c_p$ and $V_{(Y,g)}(\omega) \approx \sum_{q=1}^{Q} \omega(y_q,\vec{\gamma_q},g_p)d_q$. More details can be found in \cite{charlier2014fshape}. 
	
	The inner product metric between $X$ and $Y$ is defined as:
	\beq
	\langle V_{(X,f)}, V_{(Y,g)} \rangle_{W^*} \ = \ \sum_{p=1}^{P} \sum_{q=1}^{Q} \kappa_f(f_p,g_q) \kappa_x(x_p,y_q)\kappa_{\beta}(\vec{\beta_p},\vec{\gamma_q})c_pd_q
	\eeq
	where $\kappa_f$ and $\kappa_x$ are Gaussian kernels and $\kappa_{\beta}$ is a Cauchy-Binet kernel. This can be re-written as:
	\beq\label{eqn:Inner_fVar}
	\langle V_{(X,f)}, V_{(Y,g)} \rangle_{W^*} \, = \, \sum_{p=1}^{P} \sum_{q=1}^{Q} \! \exp\Big(\frac{-\|f_p - g_q \|^2}{\lambda_{M}^2}\Big) \exp\Big(\frac{-\|x_p - y_q\|^2}{\lambda_W^2}\Big){\Big(\frac{\vec{\beta_p}^T \vec{\gamma_q}}{c_p \, d_q} \Big)}^2 c_p \, d_q
	\eeq
	where $\lambda_M$ and $\lambda_{W}$ are kernel bandwidth parameters. For varifolds \cite{charon2013varifold}, a computational model using only fiber geometry and used for comparison in the experiments, we drop the signal values at center points. Thus, the varifolds-based representation of fibers will be: $V_X(\omega) \approx \sum_{p=1}^{P} \omega(x_p,\vec{\beta_p})c_p$ and $V_Y(\omega) \approx \sum_{q=1}^{Q} \omega(y_q,\vec{\gamma_q})d_q$. Hence, the inner product is defined as:
	\beq\label{eqn:Inner_Var}
	\langle V_X, V_Y \rangle_{W^*} \, = \, \sum_{p=1}^{P} \sum_{q=1}^{Q} \exp\Big(\frac{-\|x_p - y_q\|^2}{\lambda_W^2}\Big){\Big(\frac{\vec{\beta_p}^T\vec{\gamma_q}}{c_p \, d_q} \Big)}^2 c_p \, d_q.
	\eeq

	\vspace{-3mm}
	\subsection{Fiber Clustering using Dictionary learning and sparse coding} 
	
	For fiber clustering, we extend the dictionary learning and sparse coding based framework presented in \cite{kuldeepMiccai2015,kumar2016sparse,KUMAR2017242}. Let $V_T$ be the set of $n$ fibers modeled using functional varifolds, $\AAA \in \RR_+^{n \times m}$ be the atom matrix representing the dictionary coefficients for each fiber belonging to one of the $m$ bundles, and $\WW \in \RR_+^{m \times n}$ be the cluster membership matrix containing the sparse codes for each fiber. Instead of explicitly representing bundle prototypes, each bundle is expressed as a linear combination of all fibers. The dictionary is then defined as $\DD \ = \ V_T\AAA$. Since this operation is linear, it is defined for functional varifolds.
	
	The problem of dictionary learning using sparse coding \cite{kuldeepMiccai2015,kumar2016sparse} can be expressed as finding the matrix $\AAA$ of $m$ bundle prototypes and the fiber-to-bundle assignment matrix $\WW$ that minimize the following cost function:
	\beq\label{eqn:DL_And_SC}
	\argmin_{\AAA, \WW} \ \ \frac{1}{2} ||V_T - V_T\AAA\WW||_{W^*}^2, \quad 
	\tx{subject to:} \ \ ||\ww_i||_0 \leq S_\mr{max}.
	\eeq  
	Parameter $S_\mr{max}$ defines the maximum number of non-zero elements in $\ww_i$ (i.e., the sparsity level), and is provided by the user as input to the clustering method.
	
	An important advantage of using the above formulation is that the reconstruction error term only requires inner product between the varifolds. Let $\QQ \in \RR^{n \times n}$ be the Gram matrix denoting inner product between all pairs of training fibers, i.e., $\QQ_{ij} = \langle V_{X_i,f_i}, V_{X_j,f_j} \rangle_{W^*}$. Matrix $\QQ$ can be calculated once and stored for further computations. The problem then reduces to linear algebra operations involving matrix multiplications. The solution of Eq. (\ref{eqn:DL_And_SC}) is obtained by alternating between sparse coding and dictionary update \cite{kuldeepMiccai2015}. The sparse codes of each fiber can be updated independently by solving the following sub-problem: 
	\beq\label{eq:SparseCodeSingleFib}
	\argmin_{\ww_i \in \RR_+^{m}} \ \ \frac{1}{2} ||V_{X_i} - V_T\AAA\ww_i||_{W^*}^2, \quad 
	\tx{subject to:} \ \ ||\ww_i||_0 \leq S_\mr{max}.
	\eeq
	which can be re-written as: 
	\beq
	\argmin_{\ww_i \in \RR_+^{m}} \ \ \frac{1}{2} \big(\QQ(i,i) + \tr{\ww_i}\tr{\AAA}\QQ\AAA\ww_i - 2\QQ(i,:)\AAA\ww_i\big), \quad 
	\tx{s.t.:} \ \ ||\ww_i||_0 \leq S_\mr{max}.
	\eeq
	
	The non-negative weights $\ww_i$ can be obtained using the kernelized Orthogonal Matching Pursuit (kOMP) approach proposed in \cite{kuldeepMiccai2015}, where the most \emph{positively} correlated atom is selected at each iteration, and the sparse weights $\ww_s$ are obtained by solving a non-negative regression problem. Note that, since the size of $\ww_s$ is bounded by $S_\mr{max}$, it can be otained rapidly. Also, in case of a large number of fibers, the Nystrom method can be used for approximating the Gram matrix \cite{kumar2016sparse}. For dictionary update, $\AAA$ is recomputed by applying the following update scheme, until convergence:
	\beq
	\AAA_{ij} \ \gets \ \AAA_{ij}\dfrac {\left( \QQ \tr{\WW}\right) _{ij}} {\left( \QQ \AAA \WW \tr{\WW}\right) _{ij}}, \quad i=1,\ldots,n, \quad j=1,\ldots,m.
	\eeq
	
	\section{Experiments}
	\label{Experiments}
	
	\paragraph{Data:} 
	We evaluate different computational models on the dMRI data of 10 unrelated subjects (6 females and 4 males, age 22-35) from the Human Connectome Project (HCP) \cite{van2013wu}. DSI Studio \cite{yeh2011ntu} was used for the signal reconstruction (in MNI space, $1$mm), and streamline tracking employed to generate $50,000$ fibers per subject (minimum length $50$ mm, maximum length $300$ mm). Generalized Fractional Anisotropy (GFA), which extends standard fractional anisotropy to orientation distribution functions, was considered as along-tract measure of microstructure. While we report results obtained with GFA, any other along-tract measure may have been used. 
	
\begin{figure}[h!]
		{\centering	
			\begin{tabular}{cc}
				\includegraphics[height=3.35cm]{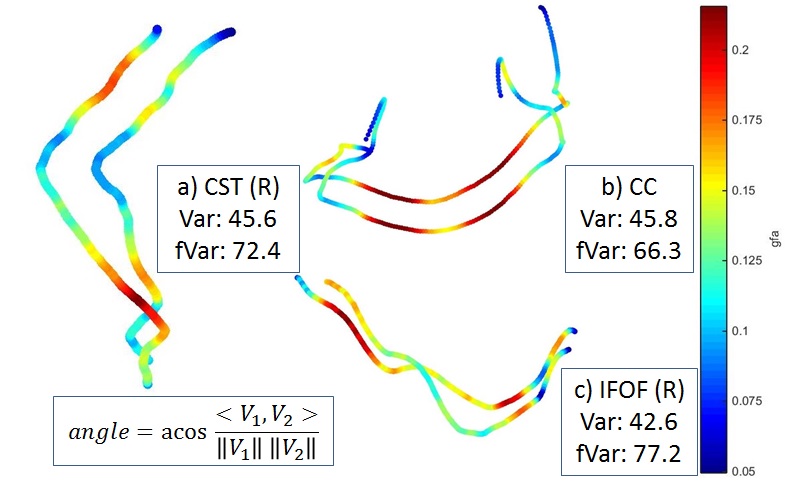} &
				\includegraphics[height=3.35cm]{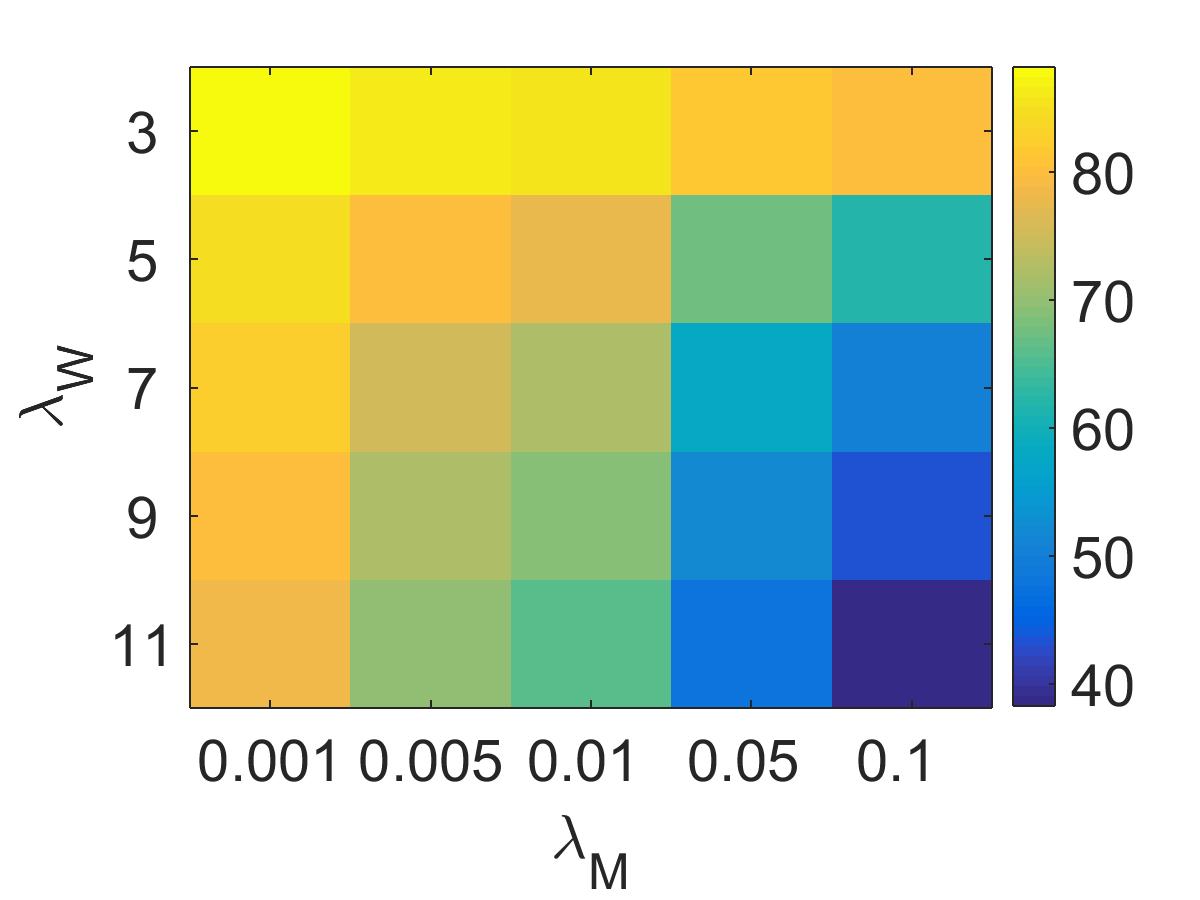} \\
				\includegraphics[height=3.35cm]{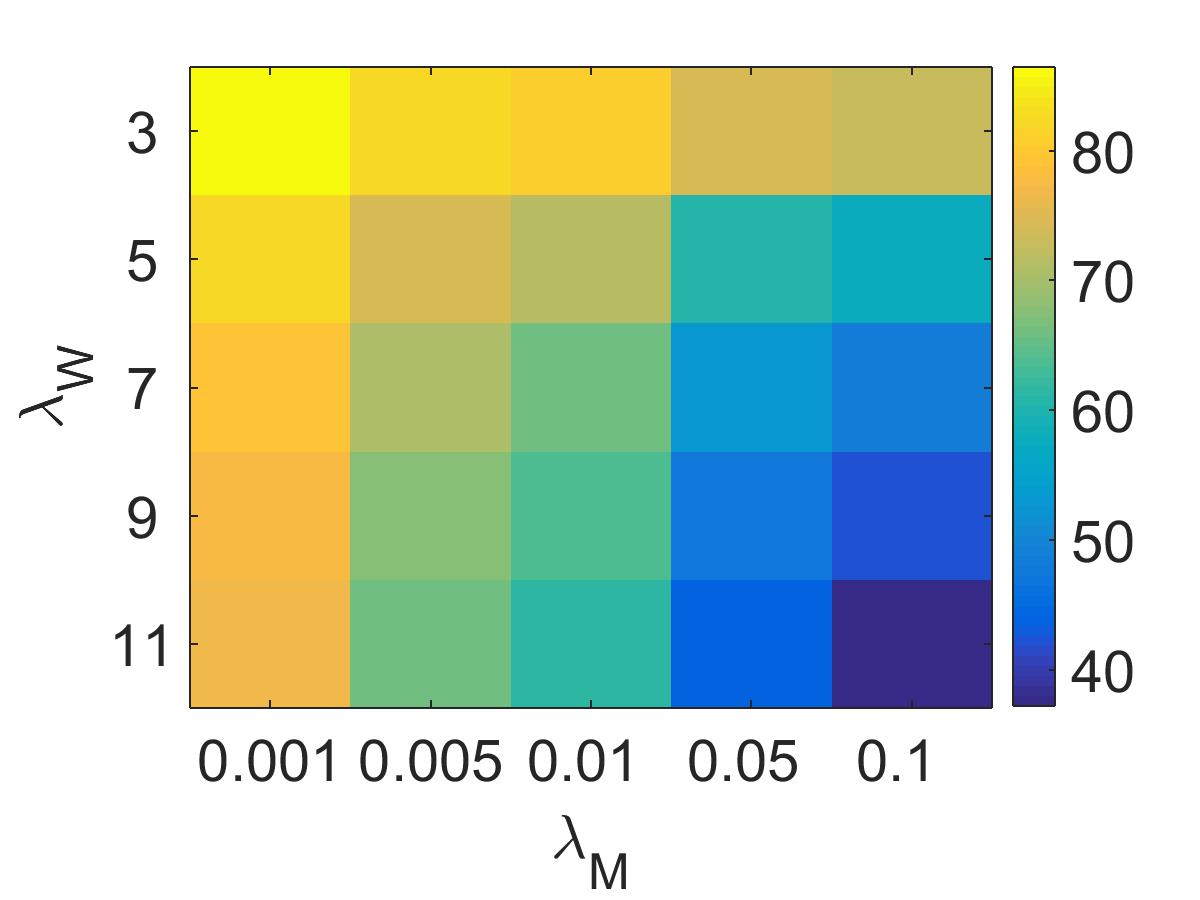} &
				\includegraphics[height=3.35cm]{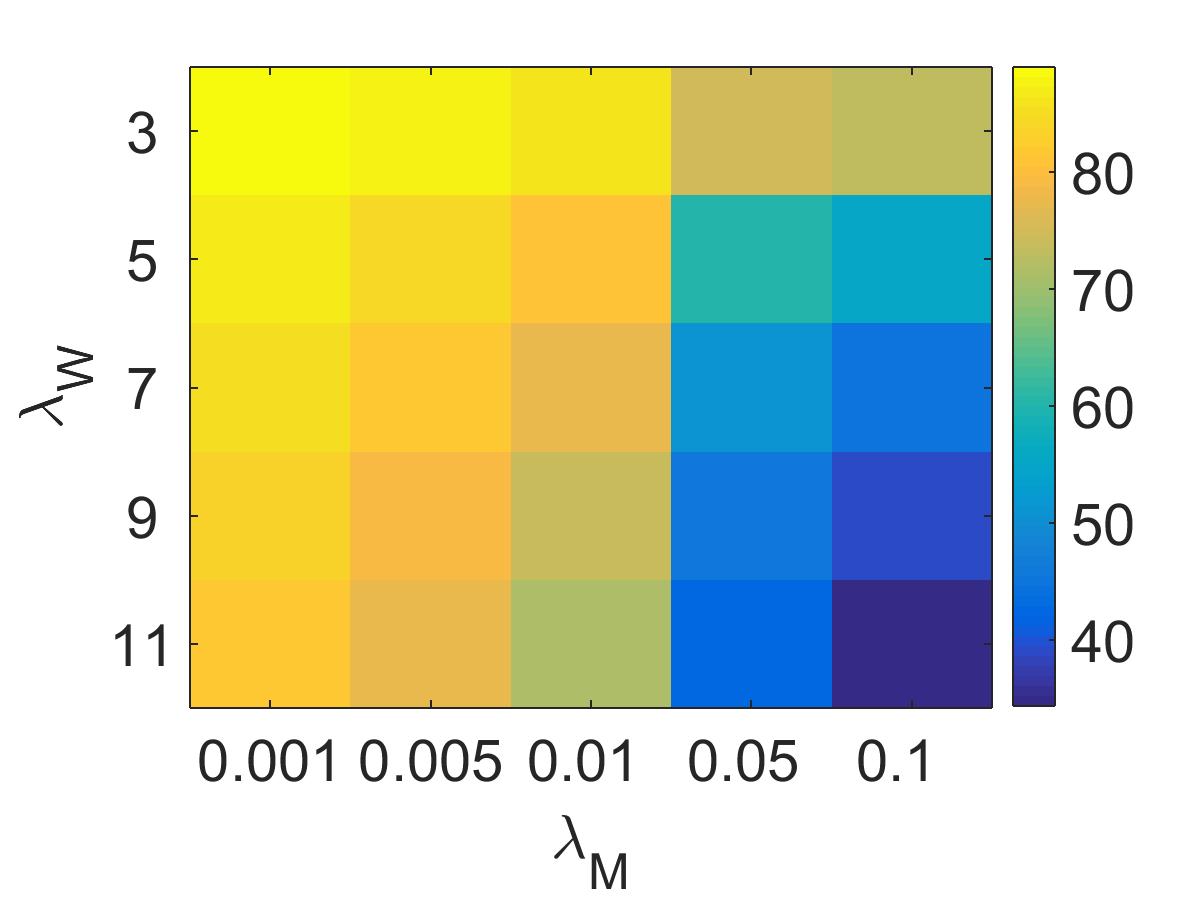}  \\
				\includegraphics[height=3.35cm]{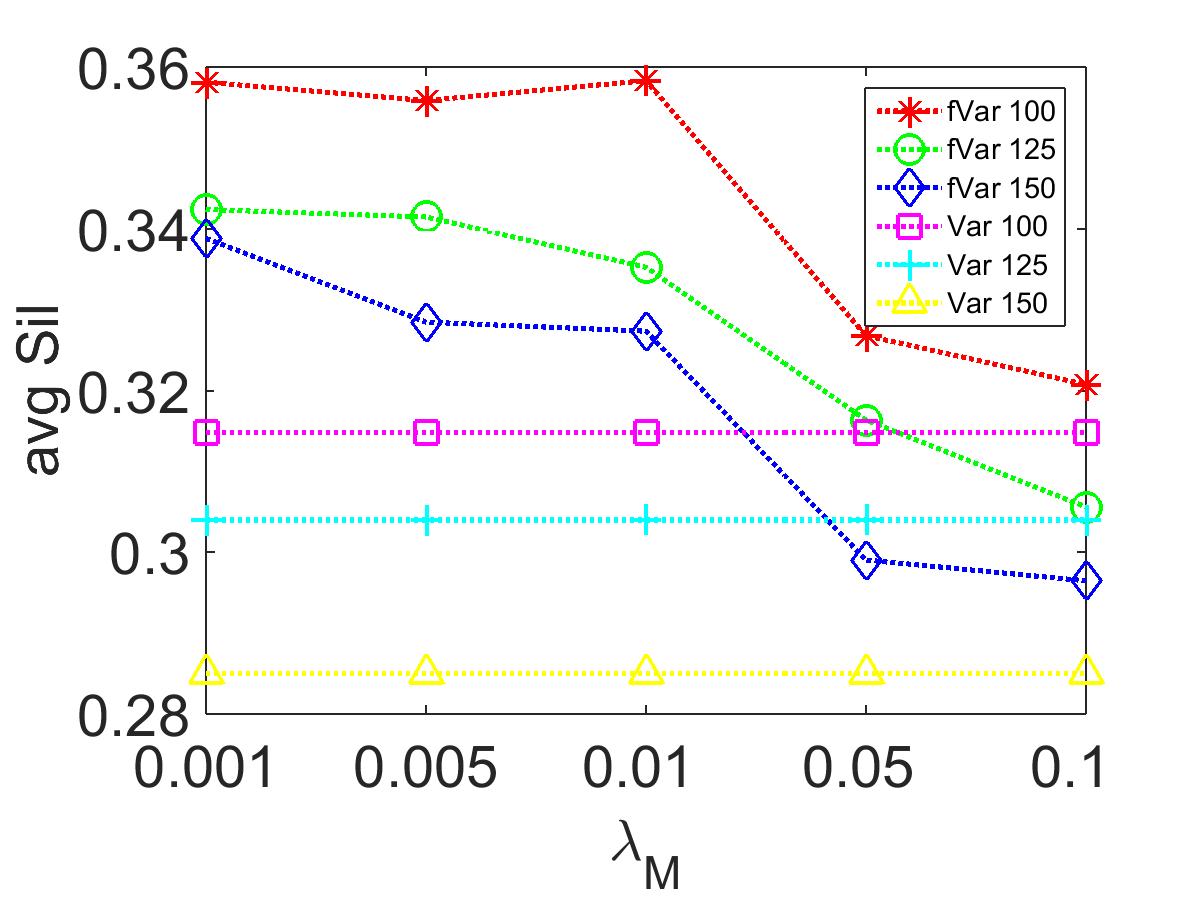} &
				\includegraphics[height=3.35cm]{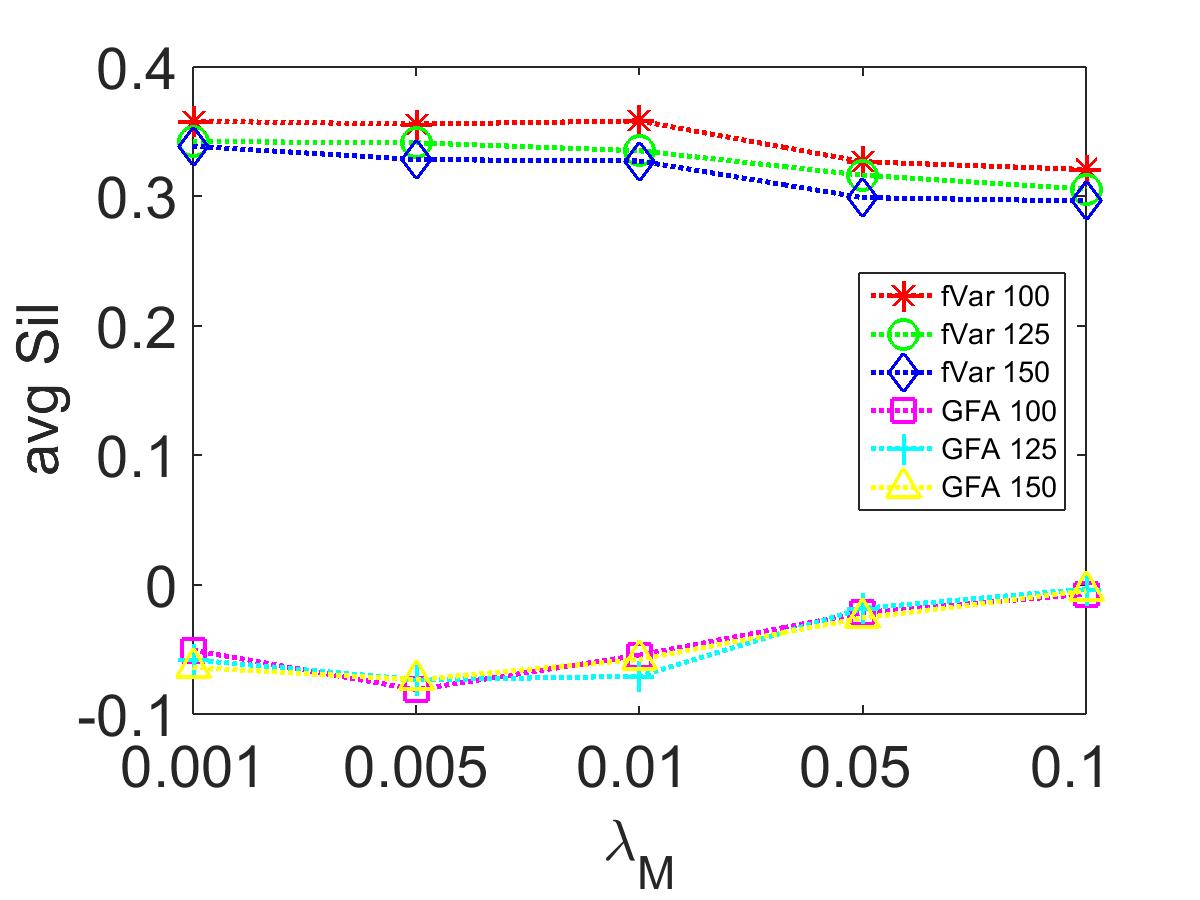}
			\end{tabular}
			\caption{Along-fiber GFA visualization and cosine similarity between pairs of fibers from three prominent bundles: a) CST (R), b) CC, c) IFOF (R), using framework of varifolds (Var) and functional varifolds (fVar) (top left), and Comparing variation of cosine similarity for the select fiber pairs over kernel bandwidth parameters $\lambda_W$ and $\lambda_M$  for the framework of functional varifolds (top right: CST (R), middle left: CC, middle right: IFOF (R)); Impact of $\lambda_{M}$ on clustering consistency (measured using Average Silhouette) for $m = 100, 125, 150$ for functional Varifolds vs Varifolds (bottom left), and functional Varifolds vs GFA only (bottom right) }
			\label{fig:Compare_Var_fVar}
		}
	\end{figure}

	\vspace{-3mm}
	\paragraph{Parameter impact:}
	We performed k-means clustering and manually selected pairs of fibers from clusters most similar to major bundles. We then modeled these fibers using different computational models, and analyzed the impact of varying the kernel bandwidth parameters. The range of these parameters were estimated by observing the values of distance between centers of fiber segments and difference between along tract GFA values for selected multiple pairs of fibers. Figure \ref{fig:Compare_Var_fVar} (top left) shows GFA color-coded fibers for $3$ pairs corresponding to a) right Corticospinal tract -- CST (R), b) Corpus Callosum -- CC, and c) right Inferior Fronto-Occipital Fasciculus -- IFOF (R). Cosine similarity (in degrees) is reported for the fiber pairs modeled using varifolds (Var) and functional varifolds (fVar), for $\lambda_W$ = 7 mm and $\lambda_M$ = 0.01. 
	
	Figure \ref{fig:Compare_Var_fVar} (top left) shows GFA color-coded fiber pairs. The color-coded visualization reflect the variation of fiber geometry, microstructure measure (i.e. GFA) along fiber, and difference in GFA along fiber for the select fiber pairs. This visualization of variation and difference in GFA values along fibers support our hypothesis that modeling along tract signal along with geometry provides additional information. The change in cosine similarity for CC from $45.8$ degrees (using varifolds) to $66.3$ degrees (using functional varifolds) while for CST (R) from $45.6$ degrees to $72.4$ degrees, reflect more drop in cosine similarity if along tract signal profiles are not similar. This shows that functional varifolds imposes penalty for different along fiber signal profiles.
	
	Figure \ref{fig:Compare_Var_fVar} also compares the impact of varying the kernel bandwidth parameters for functional varifolds using similarity angle between pairs of these selected fibers (top right: CST (R), bottom left: CC, bottom right: IFOF (R)). We show variation over $\lambda_W$ = 3, 5, 7, 9 and 11 (mm) and $\lambda_M$ = 0.001, 0.005, 0.01, 0.05, and 0.1. 
	
	\begin{figure}
		\begin{footnotesize}
			
			\begin{tabular}{C{0.5\textwidth} C{0.4\textwidth}}
				\begin{tabular}{C{13mm}C{13mm}C{13mm}C{13mm}}
					\toprule
					\textbf{Model}  & \textbf{\emph{m}=100}  & \textbf{\emph{m}=125} & \textbf{\emph{m}=150} \\
					\midrule\midrule
					fVar & \textbf{0.3624} & \textbf{0.3451} & \textbf{0.3314}  \\
					Var & 0.3356 & 0.3089 & 0.2905 \\
					GFA & -0.0579 & -0.0584 & -0.0610 \\
					MCP & 0.3240 & 0.2888 & 0.2619 \\
					\bottomrule
				\end{tabular} &
				\includegraphics[width=4cm]{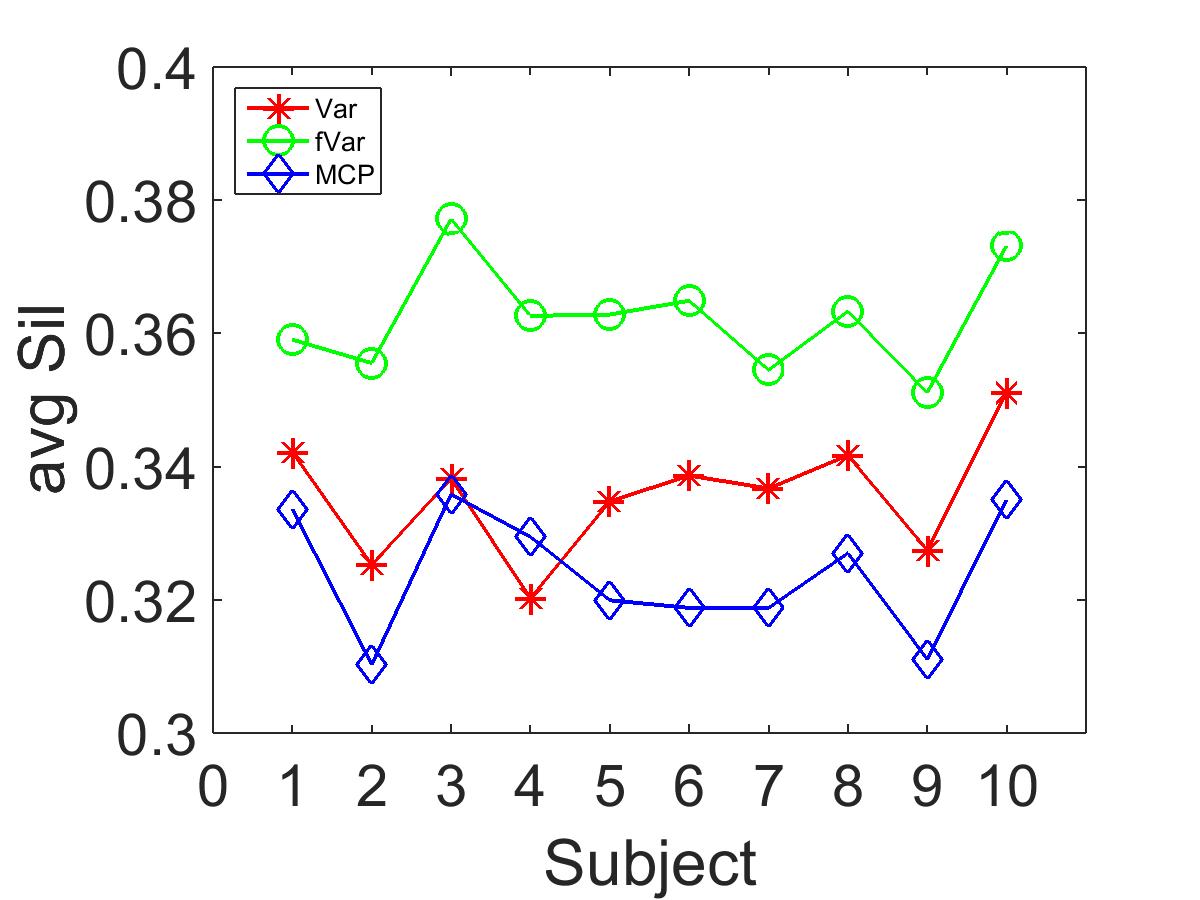}
			\end{tabular}
		\end{footnotesize}
		
		\caption{Mean silhouette obtained with Varifolds, Varifolds, GFA, and MCP, computed for varying a number of clusters, over 10 subjects and 3 seed values (\emph{left)}. Detailed results obtained for 10 subjects using $m$=100 (\emph{right}).}%
		\label{fig:Compare_Var_fVar_plot}
	\end{figure}
	\vspace{-1mm}
	
	Comparing the parameter variation images in Figure \ref{fig:Compare_Var_fVar} we observe that the cosine similarity values over the parameter space show similar trends for all $3$ pairs of fibers. This observation allows us to select a single pair of parameter values for our experiments. We have used $\lambda_W$ = $7$ mm and $\lambda_M$ = $0.01$ for our experiments based on the cosine similarity values in Figure \ref{fig:Compare_Var_fVar}. The smaller  values for $\lambda_W$ ($<7$mm) and $\lambda_M$ ($<0.01$mm) will make the current fiber pairs orthogonal while for larger values we lose the discriminative power as all fiber pairs will have very high similarity. 
	
	\vspace{-3mm}
	\paragraph{Quantitative analysis:} We report a quantitative evaluation of clusterings obtained using as functional varifolds (fVar), varifolds (var), MCP and GFA computational model. The same dictionary learning and sparse coding framework is applied for all computational models. For each of the $10$ HCP subjects, we compute the Gramian matrix using $5,000$ fibers randomly sampled over the full brain for $3$ seed values. The MCP distance $d_{ij}$ is calculated between each fiber pair $(i,j)$, as described in \cite{corouge2004towards}, and the Gramian matrix obtained using a radial basis function (RBF) kernel: $k_{ij} = \exp\big(\!-\!\gamma \!\cdot\! d_{ij}^2\big)$. Parameter $\gamma$ was set empirically to $0.007$ in our experiments. 
	
	\begin{figure}[ht]
		{\centering
			\mbox{
				\includegraphics[trim=3cm 1cm 4cm 1cm,clip=true, width=4cm,height=3.5cm,angle=90]{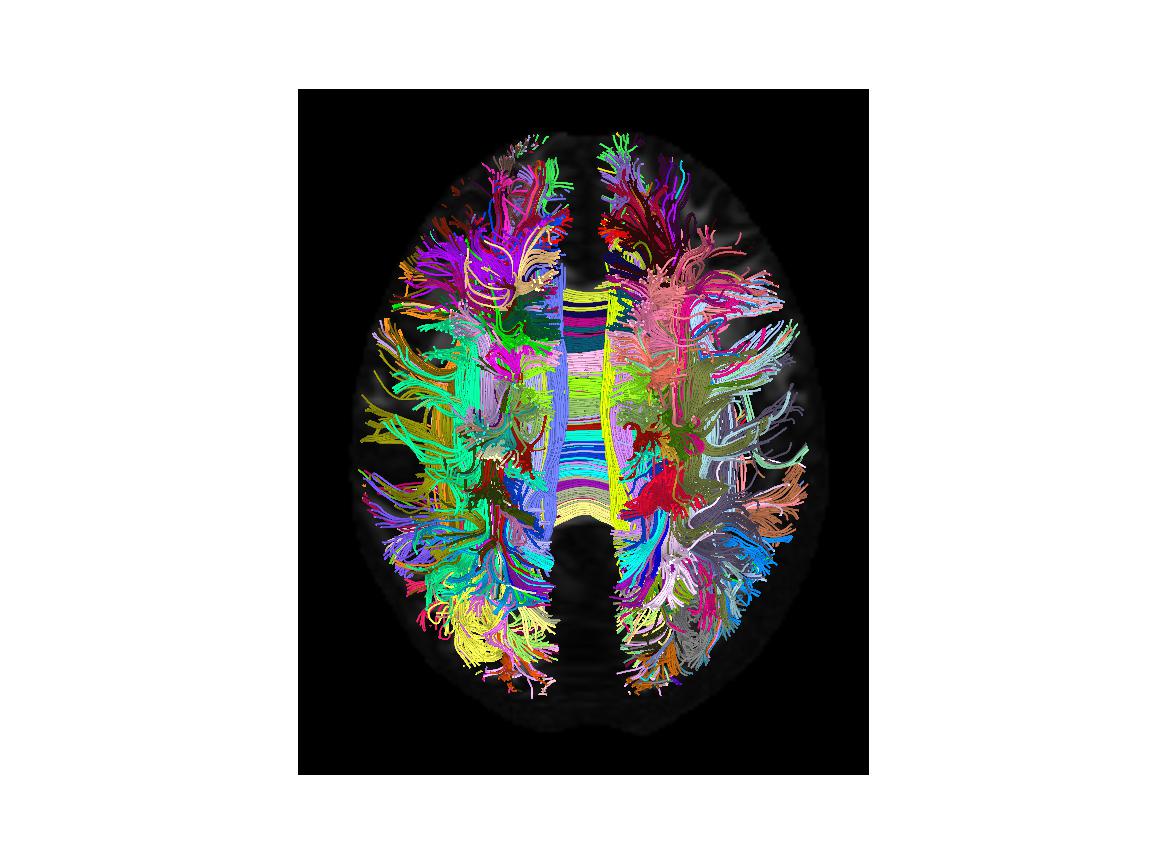}
				\hspace{1mm}
				\includegraphics[trim=3cm 1cm 4cm 1cm,clip=true, width=4cm,height=3.5cm,angle=90]{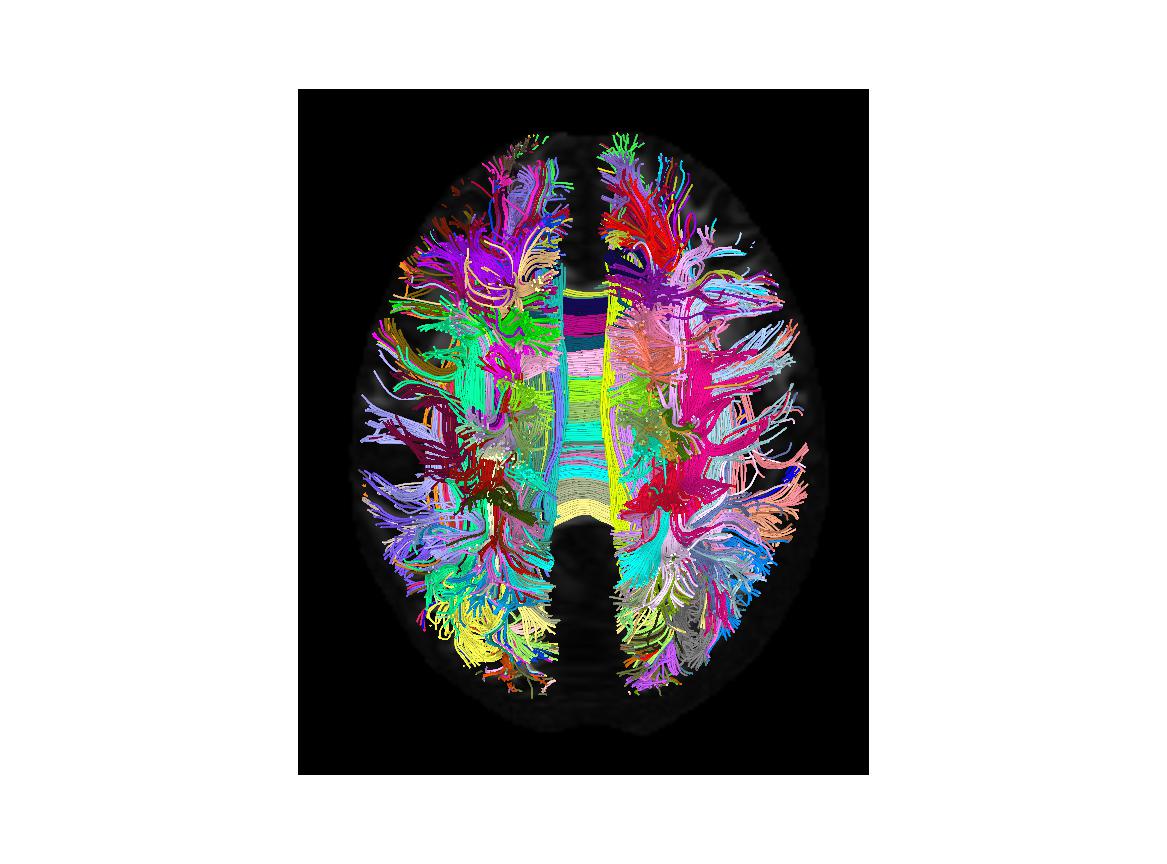}
				\hspace{1mm}
				\includegraphics[trim=3cm 1cm 4cm 1cm,clip=true, width=4cm,height=3.5cm,angle=90]{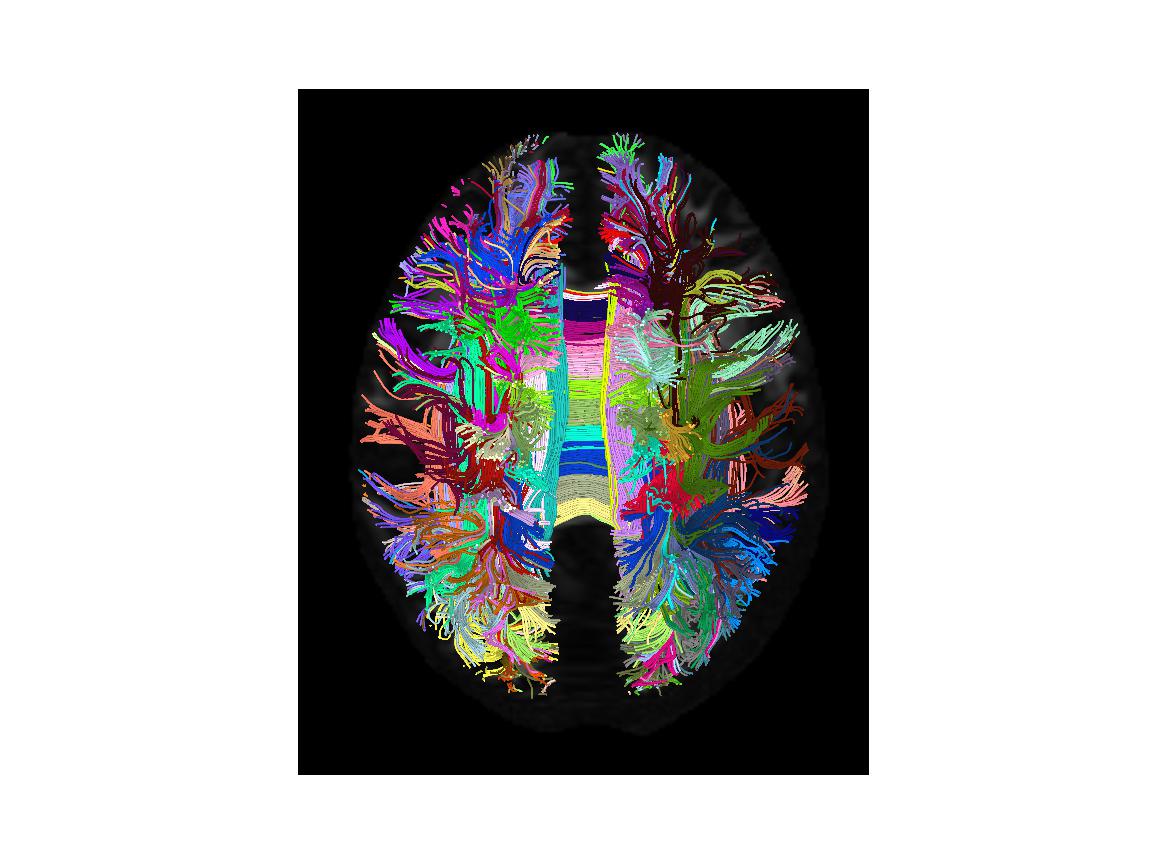}
			}
			
			\vspace{-9mm}
			
			\mbox{
				\includegraphics[trim=3cm 1cm 4cm 1cm,clip=true, width=3.5cm,height=3.5cm,angle=90]{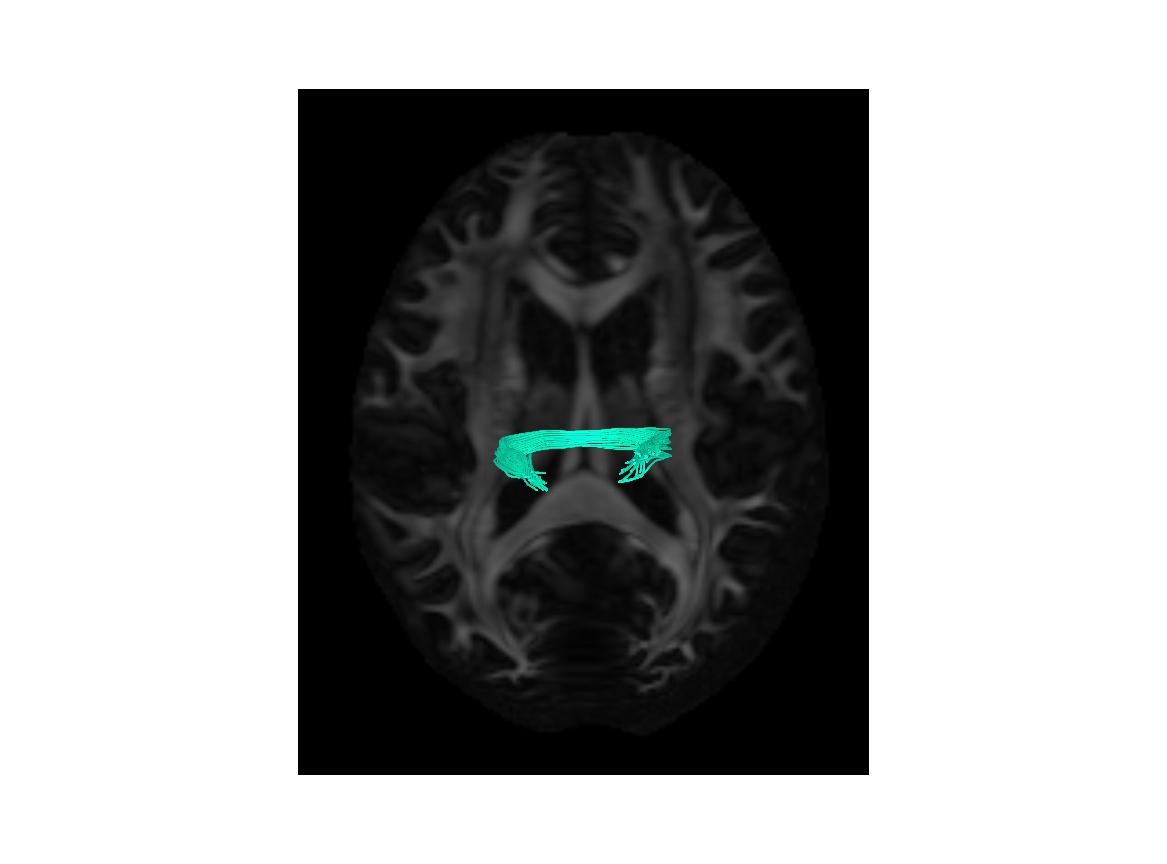}
				\hspace{1mm}
				\includegraphics[trim=3cm 1cm 4cm 1cm,clip=true, width=3.5cm,height=3.5cm,angle=90]{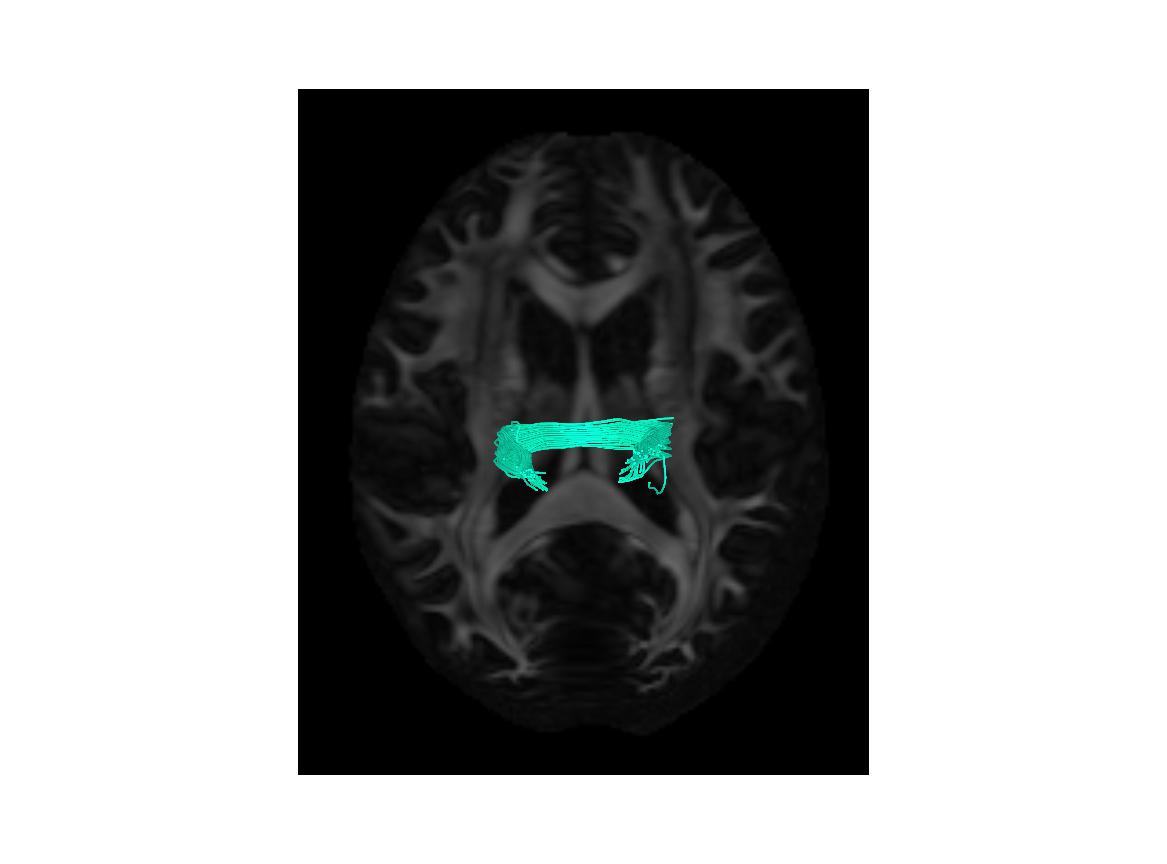}
				\hspace{1mm}
				\includegraphics[trim=3cm 1cm 4cm 1cm,clip=true, width=3.5cm,height=3.5cm,angle=90]{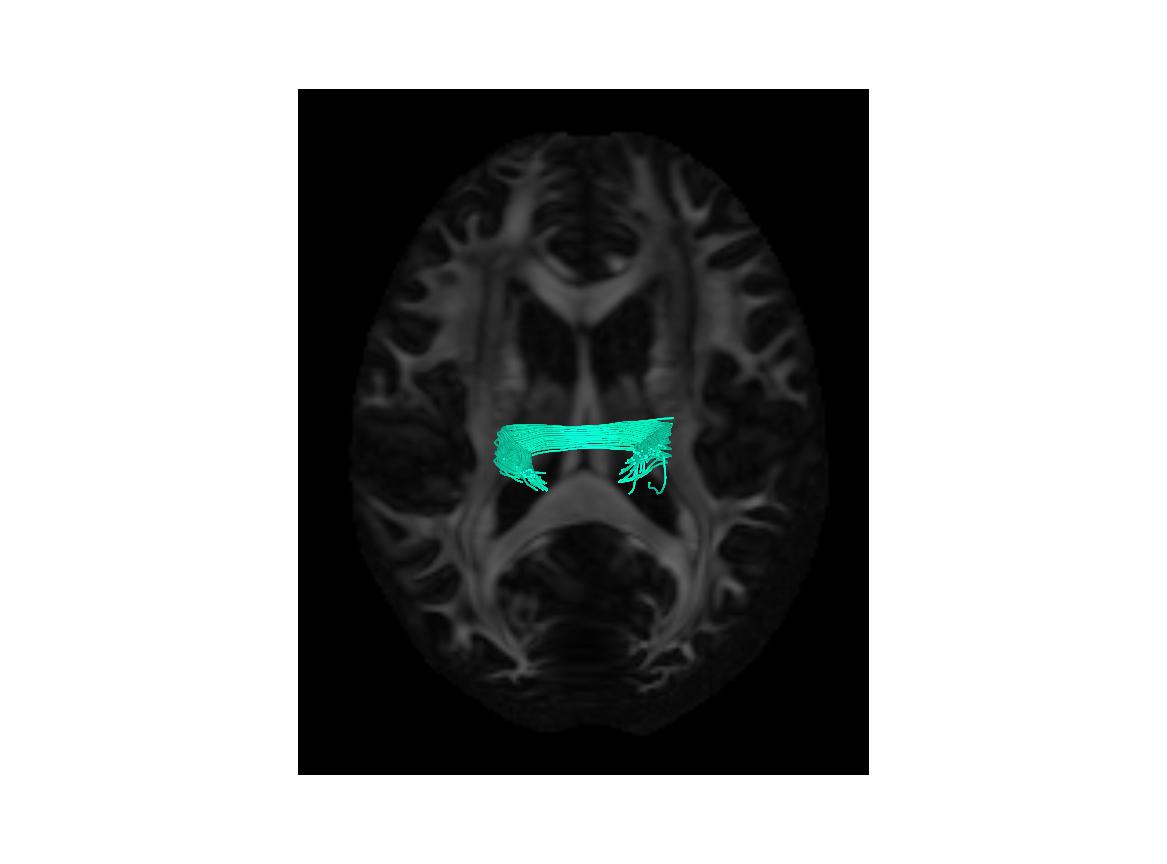}
			}	
			
			\vspace{-6mm}
			
			\mbox{		
				\hspace{-1.25mm}
				\stackunder[5pt]{		
					\includegraphics[width=3.5cm]{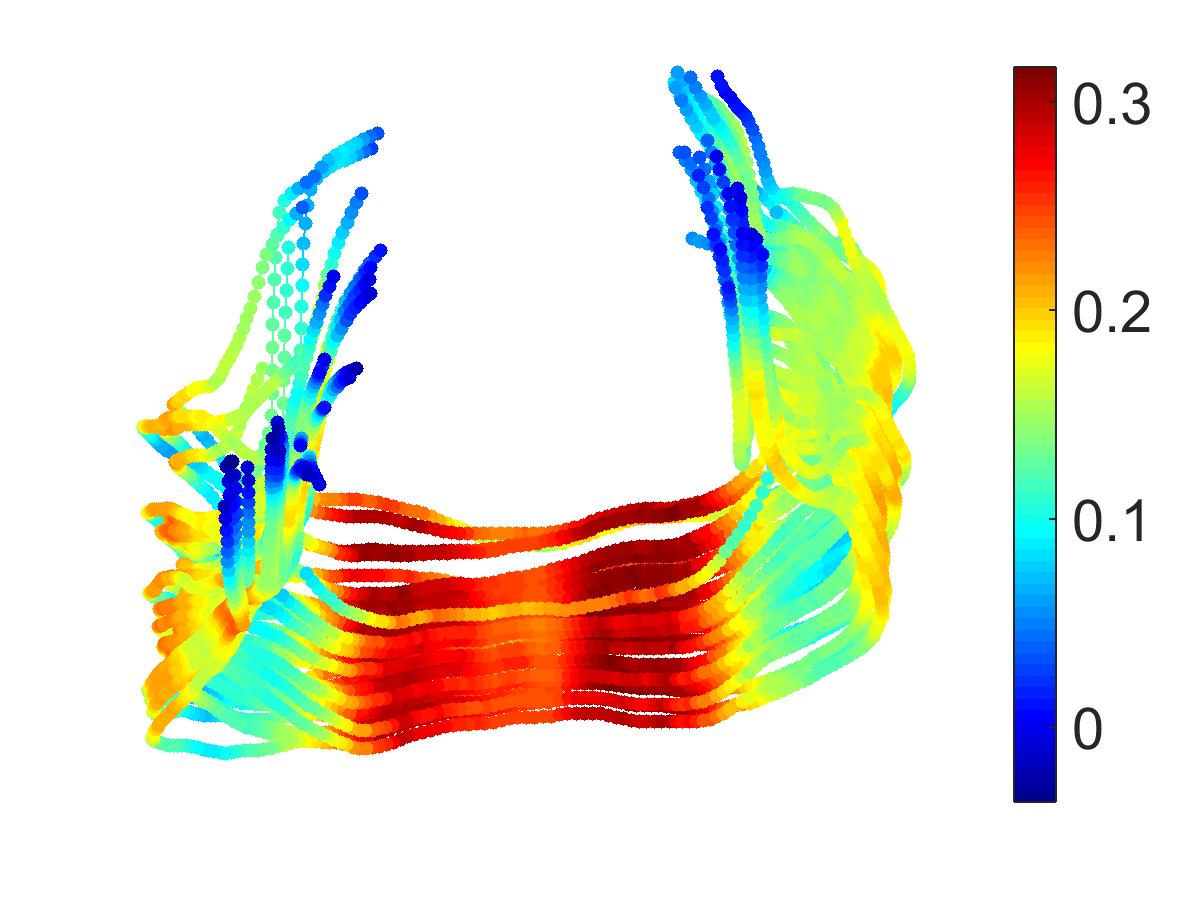}
				}{fVar}
				\stackunder[5pt]{
					\includegraphics[width=3.5cm]{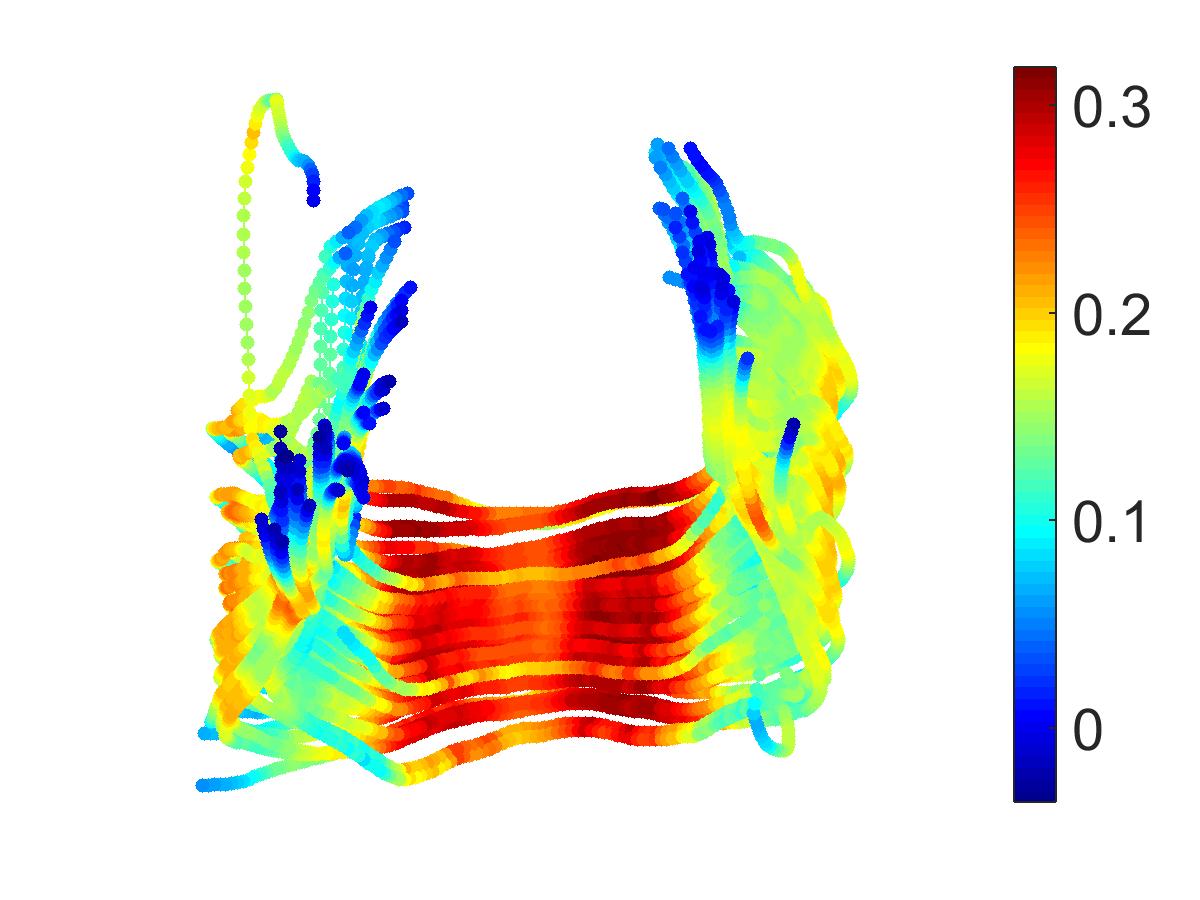}
				}{Var}
				\stackunder[5pt]{
					\includegraphics[width=3.5cm]{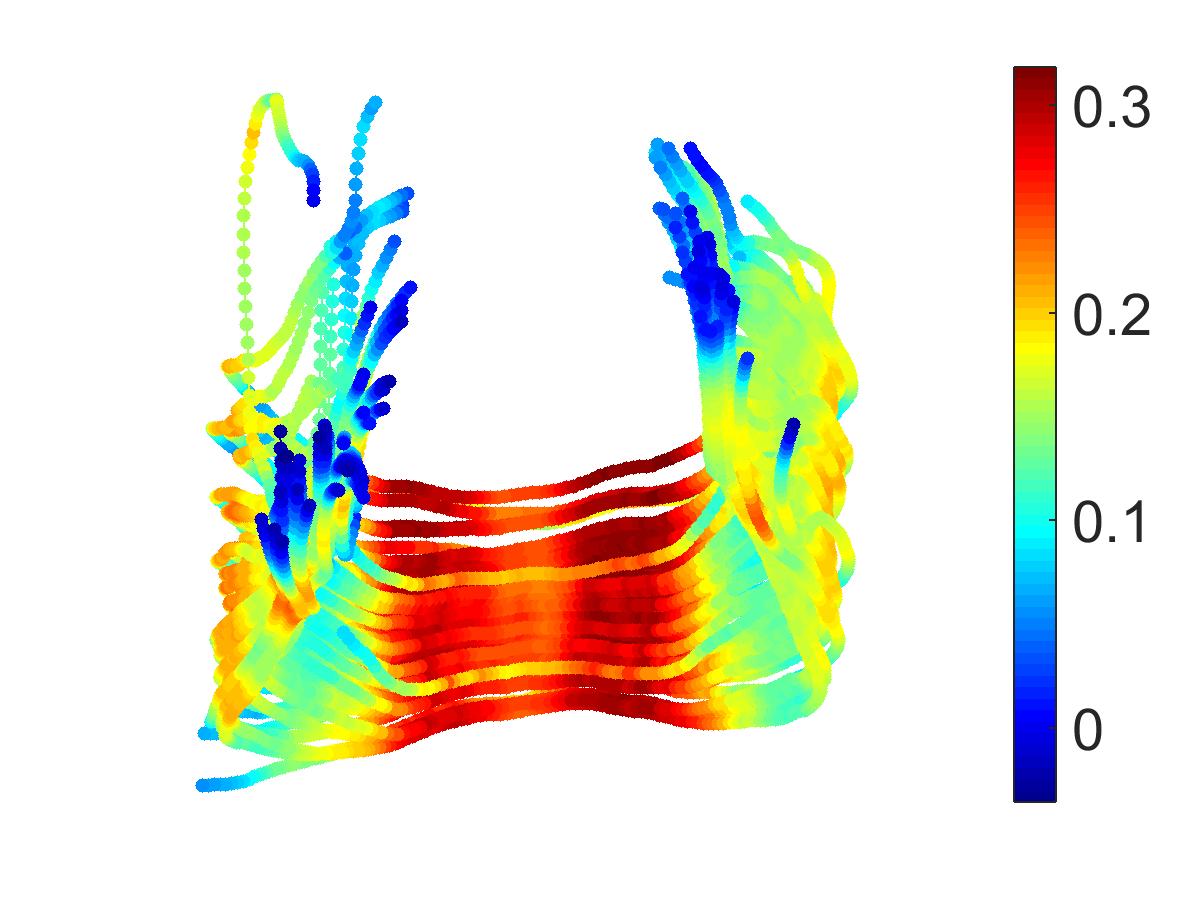}
				}{MCP}
			}
			\caption{Full clustering visualization ($m$ = 100, top row), single cluster visualization (mid row), and GFA based color coded visualization of the selected single cluster (bottom row). Using following computational models for fibers: functional varifolds (left column), varifolds (middle column), and MCP distance (right column). Superior axial views. Note: (top row) each figure has a unique color code.}
			\label{fig:Fiber_bundles}
		}
	\end{figure}
	
	Since our evaluation is performed in an unsupervised setting, we use the silhouette measure \cite{moberts2005evaluation,siless2013comparison} to assess and comparing clustering consistency. Silhouette values, which range from $-1$ to $1$, measure how similar an object is to its own cluster (cohesion) compared to other clusters (separation). Figure \ref{fig:Compare_Var_fVar} (bottom row) shows impact of $\lambda_{M}$ on clustering consistency for functional Varifolds w.r.t Varifolds and GFA only. Figure \ref{fig:Compare_Var_fVar_plot} (right) gives the average silhouette for $m$ = 100, 125, and 150 clusters, computed over $10$ subjects and $3$ seed values. The impact of using both geometry and microstructure measures along fibers is evaluated quantitatively by comparing clusterings based on functional varifolds with those obtained using only geometry (i.e., varifolds, MCP), and only along-fiber signal (i.e., GFA). As can be seen, using GFA alone leads to poor clusterings, as reflected by the negative silhouette values. Comparing functional varifolds with varifolds and GFA, we observe a consistently improved performance for different numbers of clusters. To further validate this hypothesis, we also report the average silhouette (over $3$ seed values) obtained for $10$ subjects using $m$ = 100. These results demonstrate that functional varifolds give consistently better clustering, compared to other computational models using the same framework\footnote{Silhouette analyzes only clustering consistency, not the along-fiber signal profile.}. 
	
	\vspace{-3mm}
	\paragraph{Qualitative visualization:} Figure \ref{fig:Fiber_bundles} (top row) shows the dictionary learned for a single subject ($m$ = 100) using functional varifolds (fVar), varifolds (Var), and MCP distance. For visualization purposes, each fiber is assigned to a single cluster, which is represented using a unique color. The second and third rows of the figure depict a specific cluster and its corresponding GFA color-coded profiles. We observe that all three computational models produce plausible clusterings. From the GFA profiles of the selected cluster (with correspondence across computational models), we observe that functional varifolds enforce both geometric \emph{as well as} along-tract signal profile similarity. Moreover, the clustering produced with varifolds or MCP (i.e., using only geometric properties of fibers), are similar to one another and noticeably different from that of functional varifolds. 
	
	\vspace{-4mm}
	\section{Conclusion}
	\label{Conclusions}
	
	A novel computational model, called functional varifolds, was proposed to model both geometry and microstructure measure along fibers. We considered the task of fiber clustering and integrated our functional varifolds model within framework based on dictionary learning and sparse coding. The driving hypothesis that combining along-fiber signal with fiber geometry helps tractography analysis was validated quantitatively and qualitatively using data from Human Connectome Project. Results show functional varifolds to yield more consistent clusterings than GFA, varifolds and MCP. While this study considered a fully unsupervised setting, further investigation would be required to assess whether functional varifolds augment or aid the reproducibility of results. 
	\paragraph{}
	\textbf{Acknowledgements:} Data were provided by the Human Connectome Project.
	
	
	%
	
	\vspace{-3mm}
	\bibliographystyle{splncs03}
	
	\begin{footnotesize}
		\bibliography{References_MFCA}

\begin{thebibliography}{10}
\providecommand{\url}[1]{\texttt{#1}}
\providecommand{\urlprefix}{URL }

\bibitem{charlier2014fshape}
Charlier, B., Charon, N., Trouv{\'e}, A.: The fshape framework for the
  variability analysis of functional shapes. Foundations of Computational
  Mathematics pp. 1--71 (2014)

\bibitem{charon2013varifold}
Charon, N., Trouv{\'e}, A.: The varifold representation of nonoriented shapes
  for diffeomorphic registration. SIAM Journal on Imaging Sciences  6(4),
  2547--2580 (2013)

\bibitem{colby2012along}
Colby, J.B., Soderberg, L., Lebel, C., Dinov, I.D., Thompson, P.M., Sowell,
  E.R.: Along-tract statistics allow for enhanced tractography analysis.
  Neuroimage  59(4),  3227--3242 (2012)

\bibitem{corouge2004towards}
Corouge, I., Gouttard, S., Gerig, G.: Towards a shape model of white matter
  fiber bundles using diffusion tensor {MRI}. In: ISBI 2004. pp. 344--347. IEEE
  (2004)

\bibitem{gori2014prototype}
Gori, P., Colliot, O., Marrakchi-Kacem, L., Worbe, Y., Fallani, F.D.V., Chavez,
  M., Lecomte, S., Poupon, C., Hartmann, A., Ayache, N., et~al.: A prototype
  representation to approximate white matter bundles with weighted currents.
  In: MICCAI. pp. 289--296. Springer (2014)

\bibitem{hagmann2006understanding}
Hagmann, P., Jonasson, L., Maeder, P., Thiran, J.P., Wedeen, V.J., Meuli, R.:
  Understanding diffusion mr imaging techniques: From scalar diffusion-weighted
  imaging to diffusion tensor imaging and beyond 1. Radiographics
  26(suppl\_1),  S205--S223 (2006)

\bibitem{kumar2016sparse}
Kumar, K., Desrosiers, C.: A sparse coding approach for the efficient
  representation and segmentation of white matter fibers. In: ISBI, 2016. pp.
  915--919. IEEE (2016)

\bibitem{kuldeepMiccai2015}
Kumar, K., Desrosiers, C., Siddiqi, K.: Brain fiber clustering using
  non-negative kernelized matching pursuit. In: Machine Learning in Medical
  Imaging, LNCS, vol. 9352, pp. 144--152 (2015)

\bibitem{KUMAR2017242}
Kumar, K., Desrosiers, C., Siddiqi, K., Colliot, O., Toews, M.: Fiberprint: A
  subject fingerprint based on sparse code pooling for white matter fiber
  analysis. NeuroImage  158,  242 -- 259 (2017)

\bibitem{maddah2008unified}
Maddah, M., Grimson, W.E.L., Warfield, S.K., Wells, W.M.: A unified framework
  for clustering and quantitative analysis of white matter fiber tracts.
  Medical image analysis  12(2),  191--202 (2008)

\bibitem{moberts2005evaluation}
Moberts, B., Vilanova, A., van Wijk, J.J.: Evaluation of fiber clustering
  methods for diffusion tensor imaging. In: VIS 2005. pp. 65--72. IEEE (2005)

\bibitem{o2009tract}
O'Donnell, L.J., Westin, C.F., Golby, A.J.: Tract-based morphometry for white
  matter group analysis. Neuroimage  45(3),  832--844 (2009)

\bibitem{siless2013comparison}
Siless, V., Medina, S., Varoquaux, G., Thirion, B.: A comparison of metrics and
  algorithms for fiber clustering. In: PRNI, 2013. pp. 190--193. IEEE (2013)

\bibitem{van2013wu}
Van~Essen, D.C., Smith, S.M., Barch, D.M., Behrens, T.E., Yacoub, E., Ugurbil,
  K., Consortium, W.M.H., et~al.: The wu-minn human connectome project: an
  overview. Neuroimage  80,  62--79 (2013)

\bibitem{wang2013application}
Wang, Q., Yap, P.T., Wu, G., Shen, D.: Application of neuroanatomical features
  to tractography clustering. Human brain mapping  34(9),  2089--2102 (2013)

\bibitem{wassermann2010unsupervised}
Wassermann, D., Bloy, L., Kanterakis, E., Verma, R., Deriche, R.: Unsupervised
  white matter fiber clustering and tract probability map generation:
  Applications of a {G}aussian process framework for white matter fibers.
  NeuroImage  51(1) (2010)

\bibitem{yeatman2012tract}
Yeatman, J.D., Dougherty, R.F., Myall, N.J., Wandell, B.A., Feldman, H.M.:
  Tract profiles of white matter properties: automating fiber-tract
  quantification. PloS one  7(11),  e49790 (2012)

\bibitem{yeh2011ntu}
Yeh, F.C., Tseng, W.Y.I.: Ntu-90: a high angular resolution brain atlas
  constructed by q-space diffeomorphic reconstruction. Neuroimage  58(1),
  91--99 (2011)

\end{thebibliography}
	\end{footnotesize}
	
\end{document}